# Vision-Based Road Detection in Automotive Systems:
# A Real-Time Expectation-Driven Approach

**Alberto Broggi**                                          BROGGI@CE.UNIPR.IT
**Simona Bertè**                                            SIMONA@CE.UNIPR.IT
*Dipartimento di Ingegneria dell'Informazione*
*Università di Parma*
*Viale delle Scienze*
*I-43100 Parma, Italy*

## Abstract

The main aim of this work is the development of a vision-based road detection system fast enough to cope with the difficult real-time constraints imposed by moving vehicle applications. The hardware platform, a special-purpose massively parallel system, has been chosen to minimize system production and operational costs.

This paper presents a novel approach to expectation-driven low-level image segmentation, which can be mapped naturally onto mesh-connected massively parallel SIMD architectures capable of handling hierarchical data structures. The input image is assumed to contain a distorted version of a given template; a multiresolution stretching process is used to reshape the original template in accordance with the acquired image content, minimizing a potential function. The distorted template is the process output.

## 1. Introduction

The work discussed in this paper forms part of the Eureka PROMETHEUS activities, aimed at improved road traffic safety. Since the processing of images is of fundamental importance in automotive applications, our current work has been aimed at the development of an embedded low-cost *computer vision* system. Due to the special field of application, the vision system must be able to process data and produce results in real-time. It is therefore necessary to consider data structures, processing techniques, and computer architectures capable of reducing the response time of the system as a whole.

The system considered is currently integrated on the MOB-LAB land vehicle (Adorni, Broggi, Conte, & D'Andrea, 1995). The MOBile LABoratory, the result of Italian work within the PROMETHEUS project (see Figure 1.a), comprises a camera for the acquisition and digitization of images, which pipelines data to an on-board massively parallel computer for processing. As illustrated in Figure 2, the current output configuration comprises a set of warnings to the driver, displayed by means of a set of LEDs on a control-panel (shown in Figure 1.b). But, due to the high performance levels achieved, it will be possible to replace this output device with a heads-up display showing the enhanced features superimposed onto the original image.

This paper presents a move toward the use of *top-down* control (the following feature extraction mechanism is based on a *model-driven* approach), instead of the traditional *data-driven* approach, which is generally used for data-parallel algorithms.

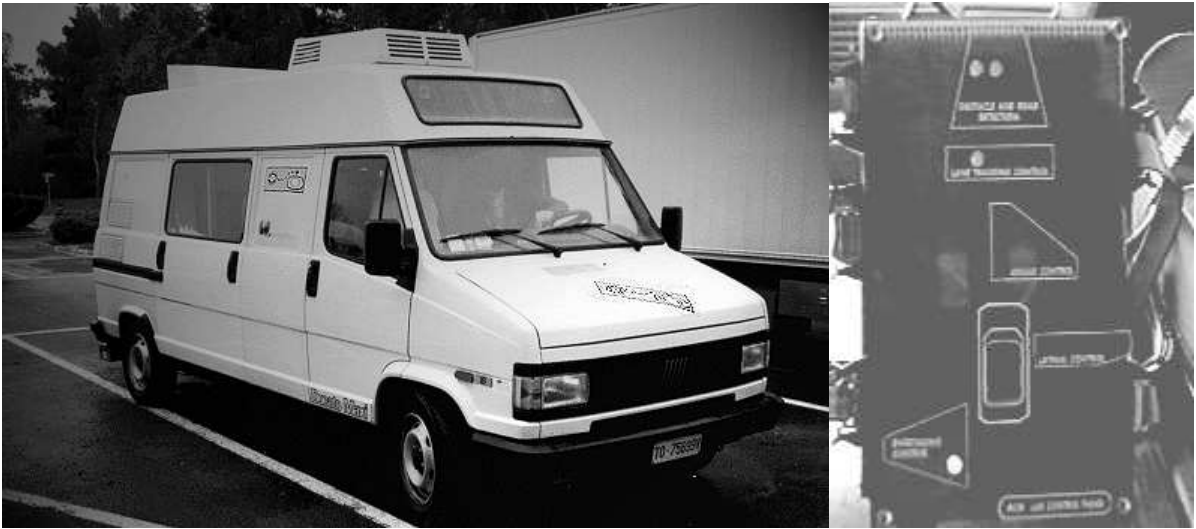

Figure 1: (a) The MOB-LAB land vehicle; (b) the control panel used as output to display the processing results

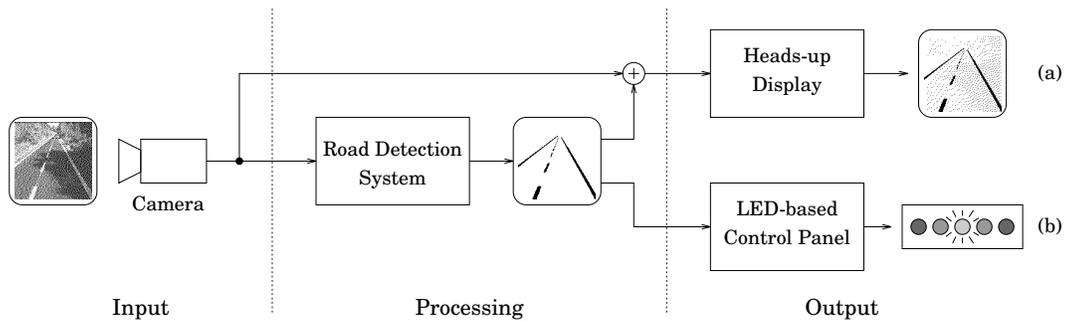

Figure 2: Block diagram of the complete system: (a) planned heads-up display output and (b) current LED-based output

Starting from the experience gained in the development of a different approach (Broggi, 1995c) based on the parallel detection of image edges pointing to the Focus of Expansion, this work presents a *model-driven* low-level processing technique aimed at road detection and enhancement of the road (or lane) image acquired from a moving vehicle. The model which contains *a-priori knowledge* of the feature to be extracted (road or lane) is encoded in the traditional data structure handled in low-level processing: a two-dimensional array. In this case, a binary image representing two different regions (road and off-road) has been chosen. Hereinafter this image will be referred to as *Synthetic Image*. It is obvious that different synthetic images must be used according to different acquisition conditions (camera position, orientation, optics, etc., which are fixed) and environment (number of lanes, one way or two way traffic, etc., which may change at run-time). In the final system implementation some *a-priori* world knowledge enables the correct synthetic model selection. As an example, Figure 3 presents several different synthetic images for different conditions.

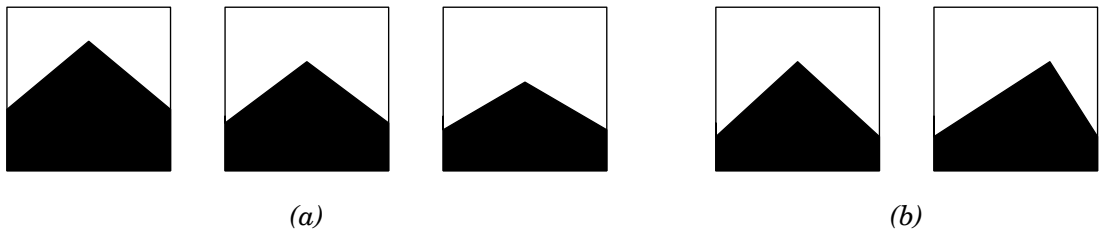

*(a)* *(b)*

Figure 3: Synthetic images used as road models for: *(a)* different camera positions and/or orientations *(b)* different number of lanes (assuming driving on the right)

The following Section presents a survey of vision-based lane detection systems; Section 3 explains the choice of the multiresolution approach; Section 4 presents the details of the complete algorithm; Section 5 discusses the performances of current implementation on the Paprica system; Section 6 presents some results and a critical analysis of the approach which leads to the present development; finally Section 7 presents some concluding remarks.

## 2. Comparison with Related Systems

Many different vision-based road detection systems have been developed worldwide, each relying on various characteristics such as different road models (two or three dimensional), acquisition devices (color or monochromatic camera, using mono or stereo vision), hardware systems (special- or general-purpose, serial or parallel), and computational techniques (template matching, neural networks, etc.).

- The Scarf system (tested on the Navlab vehicle at Carnegie Mellon University) uses two color cameras for color-based image segmentation; the different regions are classified and grouped together to form larger areas; finally a Hough-like transform is used to vote for different binary model candidates. Due to the extremely high amount of data to be processed, the two incoming color images ($480 \times 512$) are reduced to $60 \times 64$ pixel. Nevertheless, a high performance computer architecture, a 10 cell Warp (Crisman & Webb, 1991; Hamey, Web, & Wu, 1988; Annaratone, Arnould, T.Gross, H.Kung, & J.Webb, 1987), has been chosen to speed-up the processing. The system, capable of detecting even unstructured roads, reaches a processing rate of $\frac{1}{3}$ Hz (Crisman & Thorpe, 1993, 1991, 1990; Thorpe, 1989). In addition to its heavy computational load, the main problems with this approach are found in the implicit models assumed: if the road curves sharply or if it changes width, the assumed shape model becomes invalid and detection fails (Kluge & Thorpe, 1990).

- The Vits system (tested on the Alv vehicle and developed at Martin Marietta) also relies on two color cameras. It uses a combination of the red and blue color bands to segment the image, in an effort to reduce the artifacts caused by shadows. Information on vehicle motion is also used to aid the segmentation process. Tested successfully on straight, single lane roads, it runs faster than SCARF, sacrificing general capability for speed (Turk, Morgenthaler, Gremban, & Marra, 1988).

- ALVINN (tested on NAVLAB, CMU) is a neural network based $30 \times 32$ video retina designed, like SCARF, to detect unstructured roads, but it does not have any road model: it learns associations between visual patterns and steering wheel angles, without considering the road location. It has also been implemented on the Warp system, reaching a processing rate of about 10 Hz (Jochem, Pomerleau, & Thorpe, 1993; Pomerleau, 1993, 1990).

- A different neural approach has been developed at CMU and tested on NAVLAB: a $256 \times 256$ color image is segmented on a 16k processor MasPar MP-2 (MasPar Computer Corporation, 1990). A trapezoidal road model is used, but the road width is assumed to be constant throughout the sequence: this means that although the trapezoid may be skewed to the left or right, the top and bottom edges maintain a constant length. The high performance offered by such a powerful hardware platform is limited by its low I/O bandwidth; therefore a simpler reduced version (processing $128 \times 128$ images) has been implemented, working at a rate of 2.5 Hz (Jochem & Baluja, 1993).

Due to the high amount of data (2 color images) and to the complex operations involved (segmentation, clustering, Hough transform, etc.) the system discussed, even if implemented on extremely powerful hardware machines, achieve a low processing rate. Many different methods have been considered to speed-up the processing, including the processing of monochromatic images and the use of windowing techniques (Turk et al., 1988) to process only the regions of interest, thus implementing a Focus of Attention mechanism (Wolfe & Cave, 1990; Neumann & Stiehl, 1990).

- As an example, in VAMORS (developed at Universität der Bundeswehr, München) monochromatic images are processed by custom hardware, focusing only on the regions of interest (Graefe & Kuhnert, 1991). The windowing techniques are supported by strong road and vehicles models to predict features in incoming images (Dickmans & Mysliwetz, 1992). In this case, the vehicle was driven at high speeds (up to 100 kph) on German autobahns, which have constant lane width, and where the road has specific shapes: straight, constant curvature, or clothoidal. The use of a single monochromatic camera together with these simple road models allows a fast processing based on simple edge detection; a match with a structured road model is then used to discard anomalous edges. This approach is disturbed in shadow conditions, when the overall illumination changes, or when road imperfections are found (Kluge & Thorpe, 1990).

- The LANELOK system (developed at General Motors) also relies on strong road models: it estimates the location of lane boundaries with a curve fitting method (Kenue & Bajpayee, 1993; Kenue, 1991, 1990), using a quadratic equation model. In addition to being disturbed by the presence of vehicles close to the road markings, lane detection generally fails in shadow conditions. An extension for the correct interpretation of shadows has therefore been introduced (Kenue, 1994); unfortunately this technique relies on fixed brightness thresholds which is far from being a robust and general approach.

The main aim of the approach discussed in this paper, on the other hand, is to build a *low-cost* system capable of achieving *real-time* performance in the detection of *structured* roads (with painted lane markings), and robust enough to tolerate severe illumination changes such as *shadows*. Limitation to the analysis of structured environments allows the use of simple road models which, together with the processing of monocular monochromatic images on special-purpose hardware allows the achievement of low-cost high performances. The use of a high performance general-purpose architecture, such as a 10 cell Warp or a 16k MasPar MP-2 (as in the case of Carnegie Mellon's Navlab), involves high costs which are not compatible with widespread large-scale use. It is for this reason that the execution of low-level computations (efficiently performed by massively parallel systems) has usually been implemented on general purpose processors, as in the case of VaMoRs. The design and implementation of special-purpose application-oriented architectures (like Paprica, Broggi, Conte, Gregoretti, Sansoè, & Reyneri, 1995, 1994), on the other hand, keep the production costs down, while delivering very high performance levels. More generally, the features that enable the integration of this architecture on a generic vehicle are:

*(a)* its low production cost,

*(b)* its low operational cost, and

*(c)* its small physical size.

## 2.1 The Computing Architecture

Additional considerations on power consumption show that mobile computing is moving in the direction of massively parallel architectures comprising a large number of relatively slow-clocked processing elements. The power consumption of dynamic systems can be considered proportional to $CfV^2$, where $C$ represents the capacitance of the circuit, $f$ is the clock frequency, and $V$ is the voltage swing. Power can be saved in three different ways (Forman & Zahorjan, 1994), by minimizing $C$, $f$, and $V$ respectively:

- using a greater level of VLSI integration, thus reducing the capacitance $C$;
- trading computer speed (with a lower clock frequency $f$) for lower power consumption (already implemented on many portable PCs);
- reducing the supply voltage $V_{DD}$.

Recently, new technological solutions have been exploited to reduce the IC supply voltage from 5 V to 3.3 V. Unfortunately, there is a speed penalty to pay for this reduction: for a Cmos gate (Shoji, 1988), the device delay $T_d$ (following a first order approximation) is proportional to $\dfrac{V_{DD}}{(V_{DD} - V_T)^2}$, which shows that the reduction of $V_{DD}$ determines a quasi-linear increment (until the device threshold value $V_T$) of the circuit delay $T_d$. On the other hand, the reduction of $V_{DD}$ determines a quadratic reduction of the power consumption. Thus, for power saving reasons, it is desirable to operate at the lowest possible speed, but, in order to maintain the overall system performance, compensation for these increased delays is required.

The use of a lower power supply voltage has been investigated and different architectural solutions have been considered so as to overcome the undesired side effects caused by the reduction of $V_{DD}$ (Courtois, 1993; Chandrakasan, Sheng, & Brodersen, 1992). The reduction of power consumption, while maintaining computational power, can be achieved by using low cost SIMD computer architectures, comprising a large number of extremely simple and relatively slow-clocked processing elements. These systems, using slower device speeds, provide an effective mechanism for trading power consumption for silicon area, while maintaining the computational power unchanged. The 4 major drawbacks of this approach are:

- a solution based on hardware replication increases the silicon area, and thus it is not suitable for designs with extreme area constraints;
- parallelism must be accompanied by extra-routing, requiring extra-power; this issue must be carefully considered and optimized;
- the use of parallel computer architectures involves the redesigning of the algorithms with a different computational model;
- since the number of processing units must be high, if the system has size constraints the processing elements must be extremely simple, performing only simple basic operations.

This paper investigates a novel approach to real-time road following based on the use of low-cost massively parallel systems and data-parallel algorithms.

## 3. The Multiresolution Approach

The image encoding the model (*Synthetic Image*) and the image from the camera (*Natural Image*) cannot be directly compared with local computations, because the latter contains much more detail than the former. From all the known methods used to decrease the presence of details, it is necessary to choose one that does not decrease the strength of the feature to be extracted. For this purpose, a low-pass filter, such as a $3 \times 3$ neighborhood-based anisotropic average filter, would not only reduce the presence of details, but also the sharpness of the road boundaries, rendering their detection more difficult. Since the road boundaries exploit a long-distance correlation, a subsampling of both the natural and the synthetic image would lead to a comparison which was less dependent on detail content. More generally, it is much easier to detect large objects at a low resolution, where only their main characteristics are present, than at a high resolution, where the details of the specific represented object can make its detection more difficult. The complete recognition and description process, on the other hand, can only take place at high resolutions, where it is possible to detect even small details because of the preliminary results obtained at a coarse resolution.

These considerations lead to the use of a pyramidal data structure (Rosenfeld, 1984; Ballard & Brown, 1982; Tanimoto & Kilger, 1980), comprising the same image at different resolutions. Many different architectures have been developed recently to support this computational paradigm (Cantoni & Ferretti, 1993; Cantoni, Ferretti, & Savini, 1990; Fountain, 1987; Cantoni & Levialdi, 1986): where the computing architecture contains a number of processing elements which is smaller than the number of image pixels and an external processor virtualization mechanism (Broggi, 1994) is used. A useful side effect due to resolution reduction is a decrease in the number of computations to be performed. Thus, the choice of a pyramidal computational paradigm, in addition to being supported by theory, offers an advantage in terms of computational efficiency.

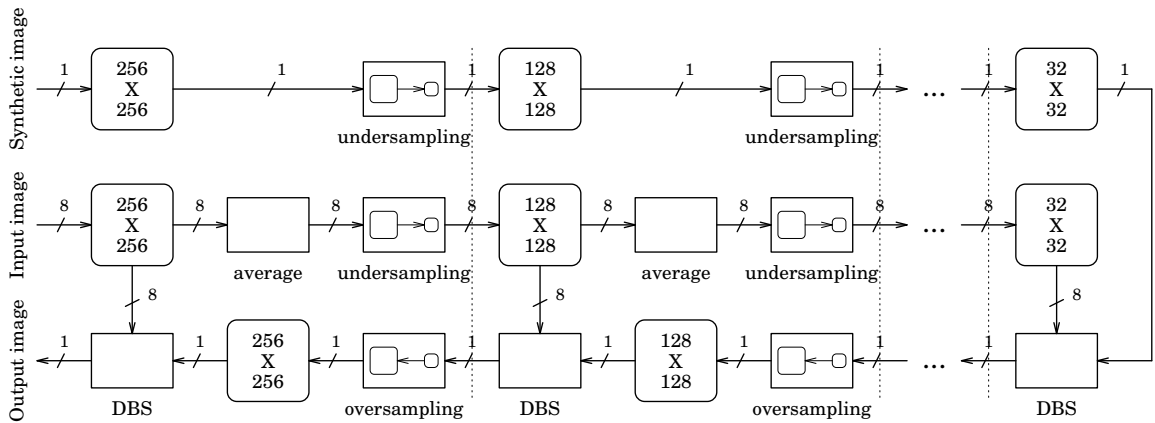

Figure 4: Block diagram of the whole algorithm; for each step the depth of every image is shown (in bit/pixel)

## 4. Algorithm Structure

As shown in Figure 4, before each subsampling the natural image is filtered. In this way it is possible to decrease both the influence of noise and redundant details, and the distortion due to *aliasing* (Pratt, 1978) introduced by the subsampling process. The image is partitioned into non-overlapping square subsets of $2 \times 2$ pixels each; the filter comprises a simple average of the pixels values for a given subset, which reduces the signal bandwidth. The set of resulting values forms the subsampled image.

The stretching of the synthetic image is performed through an iterative algorithm (*Driven Binary Stretching*, DBS), a much simpler and morphological (Haralick, Sternberg, & Zhuang, 1987; Serra, 1982) version of the "snake" technique (Blake & Yuille, 1993; Cohen & Cohen, 1993; Cohen, 1991; Kass, Witkin, & Terzopolous, 1987). The result is then oversampled, and further improved using the same DBS algorithm, until the original resolution is reached. The boundary of the stretched template represents the final result of the process.

### 4.1 The DBS Filter

The purpose of the DBS filter, illustrated in Figure 5, is to stretch the binary input model in accordance with the data encoded in the grey-tone natural image and to produce a reshaped version of the synthetic model as an output.

Usually the boundary of a generic object is represented by brightness discontinuities in the image reproducing it, therefore, in the first version, the first step of the DBS filter comprises an extremely fast and simple gradient-based filter computed on the $3 \times 3$ neighborhood of each pixel. Then, as shown in Figure 5, a threshold is applied to the gradient image, in order to keep only the most significant edges. The threshold value is now fixed, but its automatic tuning based on median filtering is currently being tested (Folli, 1994).

More precisely, two different threshold values are computed for the left and right halves of the image, in order to detect both road boundaries even under different illumination conditions.

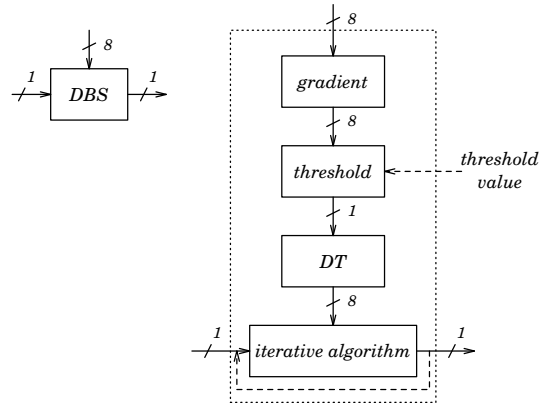

Figure 5: Block diagram of the Dbs filter

Since a 2D mesh-connected massively parallel architecture is used, an iterative algorithm must be performed in order to stretch the synthetic model toward the positions encoded in the thresholded image. A further advantage of the pyramidal approach is that the number of iterations required for successful stretching is low at a coarse resolution since the image size is small; and again, only a few iterations are required for high-resolution refinement, due to the initial coarse low-resolution stretching. Each border pixel of the synthetic image is attracted towards the position of the nearest foreground pixel of the thresholded image, as shown in Figure 6.

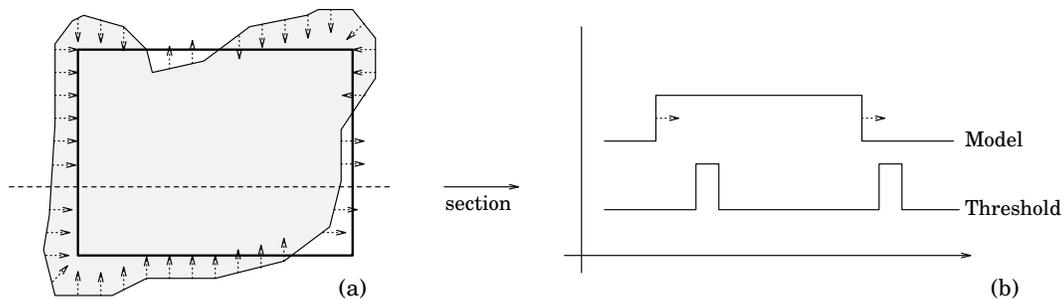

Figure 6: The attraction of the boundary pixels of the model (in grey) toward the thresholded image (the rectangular contour): (a) two-dimensional case; (b) monodimensional section

For this purpose, a scalar field $V$ is defined on $E^2$. Field $V : E^2 \to E$, links the position of each pixel $p \in E^2$ to a scalar value $V(p) \in E$, which represents the *potential* associated to the pixel itself. The set of these values is encoded in a *potential* image. The difference $V(p) - V(q)$ represents the *cost* of moving pixel $p$ toward position $q$. The scalar field is defined in such a way that a negative cost corresponds to the movement toward the nearest position of the foreground pixels $t_i$ of the thresholded image. As a consequence, the iterative process is designed explicitly to enable all the pixel movements associated to a negative cost.

Thus, the value $V(p)$ depends on the minimum distance between pixel $p$ and pixels $t_i$:

$$V(p) = -\min_i d(p, t_i) \; , \tag{1}$$

where $d : E^2 \times E^2 \to E$ represents the distance between two pixels, measured with respect to a given metric. In this case, due to the special simplicity of the implementation, a *city block* (*Manhattan distance*) metric has been chosen (see Figure 7).

| 4 | 3 | 2 | 3 | 4 |
|---|---|---|---|---|
| 3 | 2 | 1 | 2 | 3 |
| 2 | 1 | 0 | 1 | 2 |
| 3 | 2 | 1 | 2 | 3 |
| 4 | 3 | 2 | 3 | 4 |

Figure 7: *City block* or *Manhattan* distance from the central pixel

A very efficient method for the computation of the potential image using 2D mesh-connected architectures is based on the iterative application of morphological dilations (Haralick et al., 1987; Serra, 1982):

1. a scalar counter is initialized to 0; for parallel architectures which cannot run any fragment of scalar code, this counter is associated to every pixel in the image, thus constituting a "parallel" counter;

2. the counter is decremented;

3. the binary input image is dilated using a 4-connected structuring element $N$, formed by the following elements:

$$N = \Big\{ (0, 1); (0, -1); (0, 0); (1, 0); (-1, 0) \Big\} = \boxed{\begin{smallmatrix} \bullet \\ \bullet \bullet \bullet \\ \bullet \end{smallmatrix}} \quad ; \tag{2}$$

4. the value of the counter is assigned to the potential image in the positions where the pixel, due to the previous dilation, changes its state from *background* to *foreground*.

5. the process is then repeated from step 2, until the output of the morphological dilation is equal to its input.

This shows that the potential image can be generated by the application of a *Distance Transform*, DT (Borgefors, 1986) to the binary thresholded image. Due to a more efficient implementation-dependent data handling, the final version of the potential image DT is obtained by adding a constant $\lambda$ to every coefficient, so as to work with only positive values: $\lambda$ represents the maximum value allowed for grey-tone images[1]. Thus the new definition of the scalar field $V$ is:

$$V(p) = \lambda - \min_i d(p, t_i) \; . \tag{3}$$

---

1. In this specific case, since 8-bit images are considered, $\lambda = 255$.

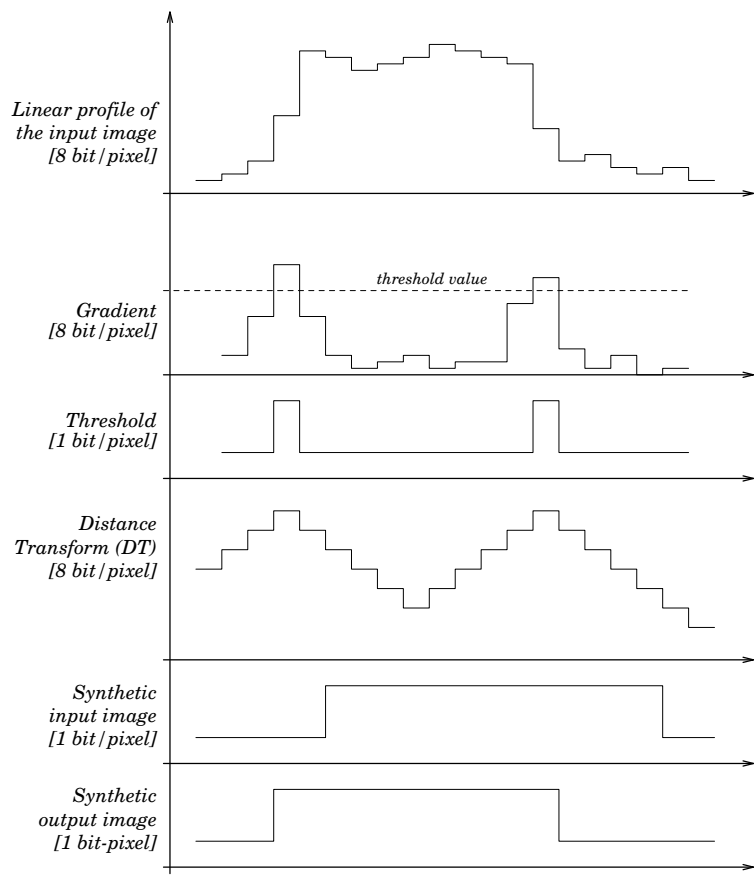

Figure 8: Example of 4 iterations of the Dbs algorithm (monodimensional case)

Furthermore, since the 'distance' information is used only by the pixels belonging to the border of the synthetic image and their neighbors, the iterative process is stopped when the DT has been computed for these pixels, producing a noticeable performance improvement.

As already mentioned, the crux of the algorithm is an iterative process whose purpose is to move the edge pixels of the synthetic image in the directions in which the DT gradient has a maximum. Figure 8 shows an example of a monodimensional stretching. As shown in Appendix A, the strength of this approach lies in the fact that the stretching algorithm can be expressed simply by a sequence of morphological operations, and can therefore be mapped efficiently on mesh-connected massively parallel architectures, achieving higher performance levels than other approaches (Cohen & Cohen, 1993; Cohen, 1991; Kass et al., 1987).

In order to describe the Dbs algorithm, let us introduce some definitions using Mathematical Morphology operators (Haralick et al., 1987; Serra, 1982) such as dilation ($\oplus$), erosion ($\ominus$), or complement ($\bar{\cdot}$). A two-dimensional binary image $S$ is represented as a subset of $E^2$, whose elements correspond to the *foreground* pixels of the image:

$$S = \left\{ s \in E^2 \,\middle\|\, s = (x, y),\ x, y \in E \right\}, \tag{4}$$

where vector $(x, y)$ represents the coordinates of the generic element $s$.

The *external edge* of $S$ is defined as the set of elements representing the difference between $S$ and the dilation of $S$ by the 4-connected structuring element $N$ shown in expression (2):

$$\mathcal{B}_e(S) = (S \oplus N) \cap \overline{S} \ . \tag{5}$$

In a similar way, the set of elements representing the difference between $S$ and its erosion using the same structuring element $N$ is defined to be the *internal edge* of $S$:

$$\mathcal{B}_i(S) = \overline{(S \ominus N)} \cap S \ . \tag{6}$$

### 4.1.1 THE ITERATIVE RULES

The set of elements $\mathcal{B}(S) = \mathcal{B}_e(S) \cup \mathcal{B}_i(S)$ are the only elements which can be inserted or removed from set $S$ by the application of a single iteration of the DBS algorithm. More precisely, two different rules are applied to the two edges: the first, applied to the external edge $\mathcal{B}_e(S)$, determines the elements to be included in set $S$; while the second, applied to the internal edge $\mathcal{B}_i(S)$, determines the elements to be removed from set $S$.

- **Rule for the external edge:**

  – each pixel of the external edge of $S$ computes the minimum value of the DT associated to its 4-connected neighbors belonging to set $S$;

  – all the pixels, whose associated DT is greater than the value previously computed, are inserted into set $S$.

  The application of this rule has the effect of expanding the synthetic image towards the foreground pixels $t_i$ which are not included in the synthetic model (see the right hand side of Figure 6.b).

- **Rule for the internal edge:**

  – each pixel of the internal edge of $S$ computes the minimum value of the DT associated to its 4-connected neighbors not belonging to set $S$;

  – all the pixels, whose associated DT is greater than the value previously computed, are removed from set $S$.

  The application of this rule has the effect of shrinking the synthetic image (see the left hand side of Figure 6.b).

Note that rule 2 is the inverse of rule 1: the latter tends to stretch the foreground onto the background, while the former acts in the opposite way, using the complement of the synthetic image.

### 4.1.2 FLAT HANDLING

Figure 8 refers to a monodimensional stretching. Unfortunately, when dealing with 2D data structures, the DT image does not present a strictly increasing or decreasing behavior, even locally. Thus, an extension to the previous rules must be considered for correct flat-handling in the 2D space.

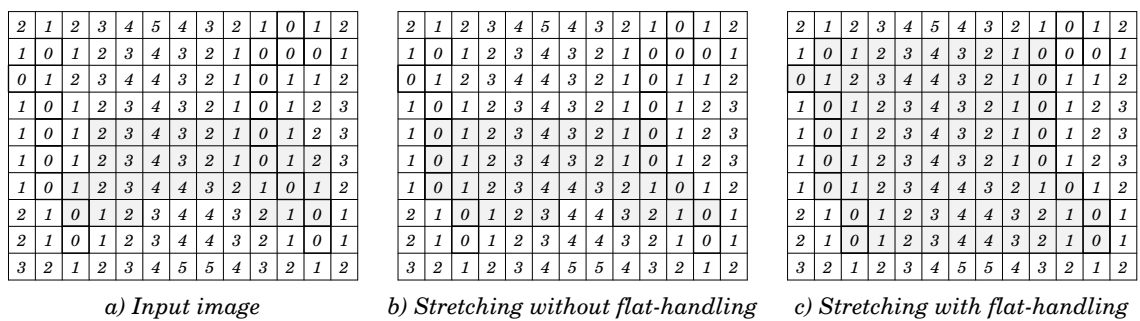

a) Input image          b) Stretching without flat-handling          c) Stretching with flat-handling

Figure 9: Two-dimensional stretching: different square markings represent different states of input binary image; dark grey areas represent the stretching area

Figure 9.a shows the modulus of the DT coefficients (in the case of $\lambda = 0$), together with the input binary image (shown in grey). The output of the iterative Dbs algorithm is presented in Figure 9.b.

Since the movement of a generic pixel towards positions holding an equal DT coefficient is expressly disabled, the resulting binary image does not completely follow the shape encoded in the DT image. A minor revision to the definition of rule 1 is thus required. Figure 9.c is obtained with the following rule applied to the external edge.

- **Rule for the external edge, including flat handling:**

  - each pixel of the external edge of $S$ computes the minimum value of the DT associated to its 4-connected neighbors belonging to set $S$;

  - all the pixels, whose associated DT is greater than the value previously computed, and all the pixels not belonging to the thresholded image, whose associated DT is equal to the value previously computed are inserted into set $S$.

The specific requirement for the pixels moving toward a flat region, of not belonging to the thresholded image, ensures that the binary image does not follow the DT chain of maxima. With such a requirement, in the specific case of Figure 9, the maxima in the upper-right hand area are not included in the resulting binary image.

## 5. Performance Analysis of the Current Implementation

Since the final aim of this work is to integrate the road detection system on a mobile vehicle, the main requirement is to achieve real-time performance. The choice of a special-purpose massively parallel architecture has already been justified in Section 1; moreover, the algorithm discussed in this paper maps naturally onto a single-bit mesh-connected Simd machine, because the whole computation can be efficiently expressed as a sequence of binary morphological operators, as shown in Appendix A (which presents the morphology-based description of the whole Dbs algorithm).

The complete processing, first tested on a Connection Machine CM-2 (Hillis, 1985), is now implemented on the special-purpose massively parallel Simd architecture Paprica.

The Paprica system (PArallel PRocessor for Image Checking and Analysis), based on a hierarchical morphology computational model, has been designed as a specialized coprocessor to be attached to a general purpose host workstation: the current implementation, consisting of a $16 \times 16$ square array of processing units, is connected to a Sparc-based workstation via a Vme bus, and installed on Mob-Lab. The current hardware board (a single 6U Vme board integrating the Processor Array, the Image and Program Memories, and a frame grabber device for the direct acquisition of images for the Paprica Image Memory) is the result of the full reengineering of the first Paprica prototype which has been extensively analyzed and tested (Gregoretti, Reyneri, Sansoè, Broggi, & Conte, 1993) for several years.

The Paprica architecture has been developed explicitly to meet the specific requirements of real-time image processing applications (Broggi, 1995c, 1995b; Adorni, Broggi, Conte, & D'Andrea, 1993); the specific processor virtualization mechanism utilized by Paprica architecture allows the handling of pyramidal data structures without any additional overhead equipment.

As shown in Figure 2, the output device can be:

(a) a heads-up display in which the road (or lane) boundaries are highlighted;

(b) a set of Leds indicating the relative position of the road (or lane) and the vehicle.

Starting from $256 \times 256$ grey-tone images and after the resolution reduction process, in the first case (a) the initial resolution must be recovered in order to superimpose the result onto the original image. In the second case (b), on the other hand, due to the high quantization of the output device, the processing can be stopped at a low resolution (e.g., $64 \times 64$), where only a stripe of the resulting image is analyzed to drive the Leds. Table 1 presents the computational time required by each step of the algorithm (considering 5 DT iterations and 5 Dbs iterations for each pyramid level).

| Operation | Image size | Time [ms] |
|---|---|---|
| Resolution Reduction | $256^2 \rightarrow 32^2$ | 149.70 |
| DT, Dbs & Oversampling | $32^2 \rightarrow 64^2$ | 18.55 |
| DT, Dbs & Oversampling | $64^2 \rightarrow 128^2$ | 69.13 |
| DT, Dbs & Oversampling | $128^2 \rightarrow 256^2$ | 296.87 |
| Total (Led output) | $256^2 \rightarrow 32^2 \rightarrow 64^2$ | 168.25 |
| Total (heads-up display output) | $256^2 \rightarrow 32^2 \rightarrow 256^2$ | 534.25 |

Table 1: Performance of Paprica system; the numbers refer to 5 iterations of the DT process and 5 iterations of the Dbs filter for each pyramid level.

In the current Mob-Lab configuration, the output device (shown in Figure 1.b) consists of a set of 5 Leds: Table 1 shows that the $256^2 \rightarrow 32^2 \rightarrow 64^2$ filtering of a single frame takes $150 \div 180$ ms (depending on the number of iterations required), allowing the acquisition and processing of about 6 frames per second.

Moreover, due to the high correlation between two consecutive sequence frames, the final stretched template can be used as the input model to be stretched by the processing of the following frame. In these conditions a lower number of Dbs and DT iterations are needed for the complete template reshaping, thus producing a noticeable performance improvement. The reduction in the time required to process a single frame also increases the correlation between the current and the following frame in the sequence, thus allowing a further reduction in computation time[2].

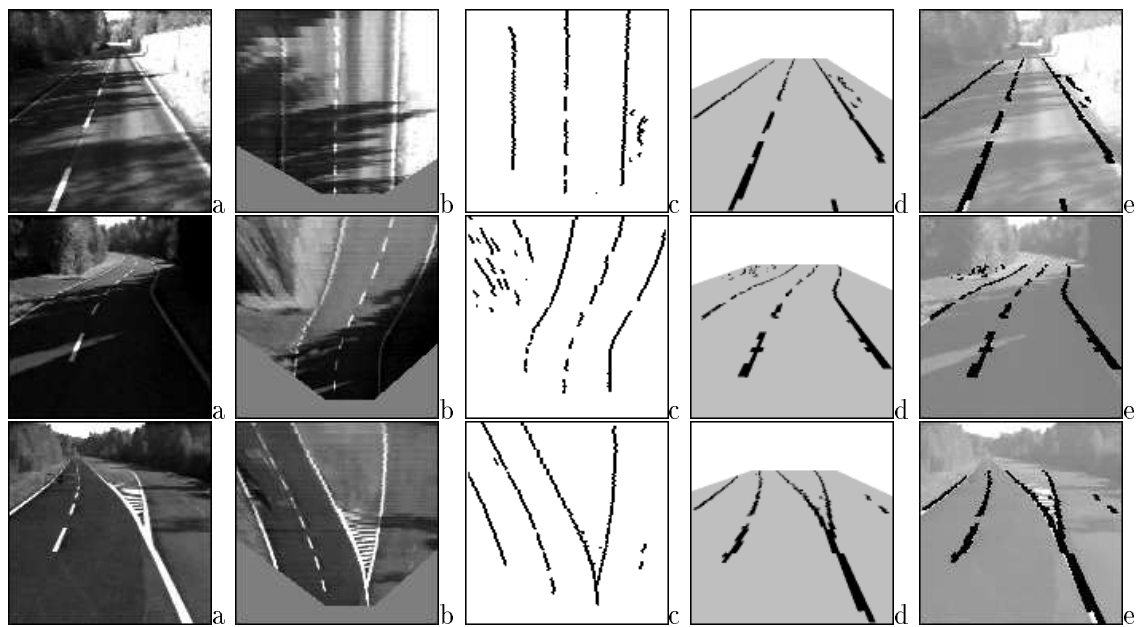

Figure 10: The detection of road markings through the removal of the perspective effect in three different conditions: straight road with shadows, curved road with shadows, junction. (a) input image; (b) reorganized image, obtained by non-uniform resampling of (a); (c) result of the line-wise detection of black-white-black transitions in the horizontal direction; (d) reintroduction of the perspective effect, where the grey areas represent the portion of the image shown in (c); (e) superimposition of (d) onto a brighter version of the original image (a).

Images acquired in many different conditions and environments have been used for extensive experimentation, which was performed first off-line on a functional simulator implemented on an 8k processor Connection Machine CM-2 (Hillis, 1985) and then in real-time on the PAPRICA hardware itself on MOB-LAB. The complete system was demonstrated at the final PROMETHEUS project meeting in Paris, in October 1994: the MOB-LAB land vehicle was driven around the two-lane track at Mortefontaine, under different conditions (straight and curved roads, with shadows and changing illumination conditions and with other vehicles on the path).

---

2. In the processing of highly correlated sequences (namely when the road conditions change slowly or in an off-line slow-motion tape playback in the laboratory) the computational time required by the processing of a single frame can reach a minimum value of 150 ms.

The performances obtained during this demonstration allowed the main limitations of the system to be detected and enabled a critical analysis of the approach, thus leading to proposals for its development.

## 6. Critical Analysis and Evolution

The approach discussed achieves good performances in terms of output quality when the model matches (or is sufficiently similar to) the road conditions, namely when road markings are painted on the road surface (on structured roads, see Figures 11.a and 12.a) inducing a sufficiently high luminance gradient. This approach is successful when the road or lane boundaries can be extracted from the input image through a gradient thresholding operation (see Figures 11.b and 12.b). Unfortunately, this is not always possible, for example when the road region is a patch of shadow or sunlight, as in Figure 13.a. In this case the computation of the DT starting from the thresholded image (see Figure 13.b) is no longer significant: a different method must be devised for the determination of the binary image to be used as input for the Distance Transform.

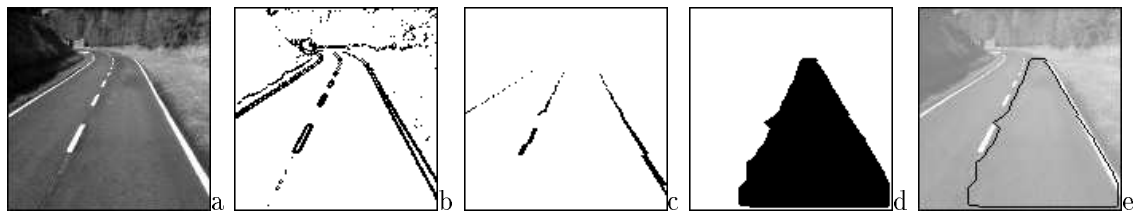

Figure 11: Lane detection on a straight road: (a) input image; (b) image obtained by thresholding the gradient image; (c) image obtained by the perspective-based filter; (d) stretched template; (e) superimposition of the edges of the stretched template onto the original image. In this case both the thresholded gradient (b) and the perspective-based filtered (c) images can be used as input to the DT.

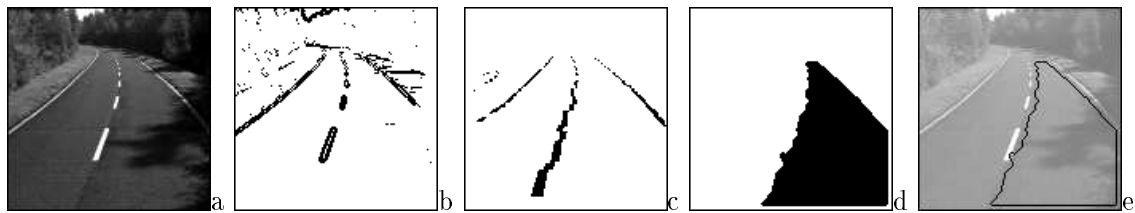

Figure 12: Lane detection on a curved road: also in this case both the thresholded gradient (b) and the perspective-based filtered (c) images can be used as input to the DT.

In a recent work (Broggi, 1995a) an approach based on the removal of the perspective effect is presented and its performances discussed. A transform, a non-uniform resampling similar to what happens in the human visual system (Zavidovique & Fiorini, 1994; Wolfe &

Cave, 1990), is applied to the input image (Figures 10.a); assuming a flat road, every pixel of the resampled image (Figures 10.b) now represent the same portion of the road[3]. Due to their constant width within the overall image, the road markings can now be easily enhanced and extracted by extremely simple morphological filters (Broggi, 1995a) (Figures 10.c). Finally the perspective effect is reintroduced (Figures 10.d).

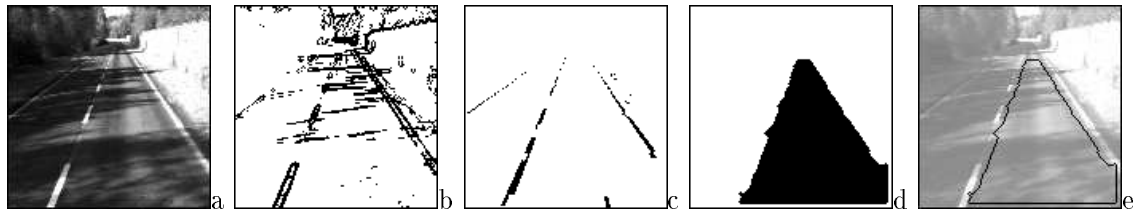

Figure 13: Lane detection on a straight road with shadows: this is a case where the thresholded gradient (b) cannot be used to determine the DT image due to the noise caused by shadows. On the other hand, the perspective-based processing is able to filter out the shadows and extract the road markings: the DT is in fact determined using image (c) instead of (b).

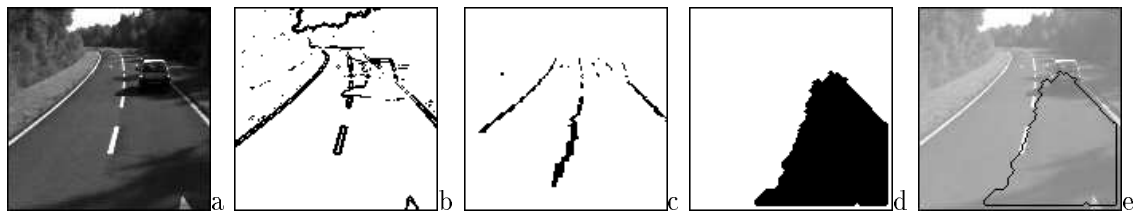

Figure 14: Lane detection on a straight road with a vehicle in the path: this is the only case in which the Dbs algorithm cannot stretch the binary model successfully, due to the presence of the vehicle occluding the road markings. The use of a pair of stereo cameras to remove this problem is currently being investigated.

Due to the high effectiveness of this filter, in the test version of the system currently operational in the laboratory, the Dbs filter has been improved by replacing the gradient thresholding with the perspective-based filter. The *extended* version of the Dbs filter is shown in Figure 15.

Since the removal (and reintroduction) of the perspective effect can be reduced to a mere image resampling and the filter is based on simple morphological operators (Broggi, 1995a), the implementation on the Paprica system is straightforward. The preliminary results obtained by the current test version of the system are encouraging both for the output quality (the problems caused by shadows are now resolved) and for computation time: a single frame is processed in less than 100 ms, thus allowing the processing of about 10 frames per second. The improvement of the DBS process by means of the perspective-based filter, in addition to allowing correct road (or lane) detection in presence of shadows, can be implemented extremely efficiently on the PAPRICA system, taking advantage of a specific hardware extension designed explicitly for this purpose (non uniform-resampling) (Broggi, 1995a).

---

3. Note that, due to the perspective effect, the pixels in the lower part of the original image (Figures 10.a) represent a few cm$^2$ of the road region, while the central pixels represent a few tens of cm$^2$. The non-uniform resampling of the original image has been explicitly designed for a homogeneous distribution of the information among all the pixels in the image.

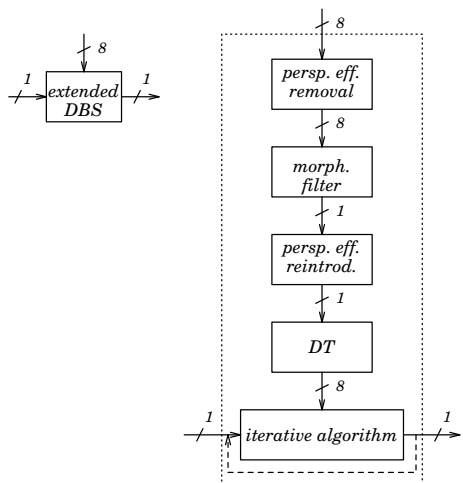

Figure 15: Diagram of the extended DBS filter including the perspective-based filtering

Figures 11, 12, 13, and 14 show the results of the processing in different conditions: straight and curved road, with shadows and other vehicles in the path, respectively[4]. In the last case the lane cannot be detected successfully due to the presence of an obstacle

The use of a pair of stereo images is currently being investigated to overcome this problem: the removal of the perspective effect from both the stereo images would lead to the same image *iff* the road is flat, namely if no obstacles are found on the vehicle path. A difference in the two reorganized images (namely when an obstacle is detected) would cause the algorithm to stop the lane detection and warn the driver.

The general nature of the presented approach enables the detection of other sufficiently large-sized features: for example, using different synthetic models it is possible to detect road or lane boundaries, as shown in Figure 16.

## 7. Conclusions

In this paper a novel approach for the detection of road (or lane) boundaries for vision-based systems has been presented. The multiresolution approach, together with top-down control, allows achievement of remarkable performances in terms of both computation time (when mapped on massively parallel architectures) and output quality.

---

4. Some sequences in MPEG format (200 and 1000 frames respectively) are available via World Wide Web at `http://WWW.CE.UniPR.IT/computer_vision`, showing lane detection in particularly challenging conditions. The images shown in Figures 11, 12, 13, 14, and 16 are part of the sequences available in the previously mentioned WWW page.

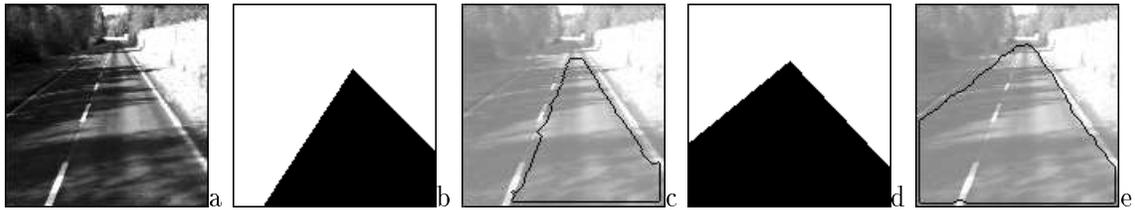

Figure 16: (a) input image; (b) synthetic model used for *lane* detection; (c) superimposition of the edges of the stretched model onto the original input image; (d) synthetic model used for *road* detection; (e) superimposition of the edged of the stretched model onto the original input image.

A perspective-based filter has been introduced to improve system performance in shadow conditions. However, even if the perspective-based filter alone is able to extract the road markings with a high degree of confidence, the hierarchical application of the DBS filter (and its extension to the handling of image sequences) is of basic importance since it allows the exploitation of the temporal correlation between successive sequence frames (performing solution tracking).

The presence of an obstacle in the vehicle path is still an open problem, which is currently being approached using stereo vision (Broggi, 1995a).

The algorithm has been implemented on the Paprica system, a massively parallel low-cost Simd architecture; because of its specific hardware features, Paprica is capable of processing about 10 frames per second.

## Appendix A. The Morphological Implementation of the DBS Filter

In this appendix, the rule for the external edge will be considered, assuming step $n$ in the iterative process. Recalling the Mathematical Morphology notations used to identify a grey-tone two-dimensional image, the DT image is a subset of $E^3$:

$$DT = \{d \in E^3 \mid d = (u, v), \ v = V(u), \ \forall \ u \in E^2\} \ , \tag{7}$$

where $u$ represents the position of element $d$ in $E^2$, and $v$ represents its *value*.

The *pixelwise masking* operation between a binary and a grey-tone image is here defined as a function

$$\odot : E^2 \times E^3 \rightarrow E^3 \tag{8}$$

such that $A \odot B$ represents a subset of $B$ containing only the elements $b = (u, v)$ whose position vector $u \in E^2$ also belongs to $A$:

$$A \odot B \stackrel{\Delta}{=} \{x \in E^3 \mid x = (u, v) \in B, \ u \in A\} \tag{9}$$

In order to compute the minimum value of the $DT$ in the specified neighborhood, let us consider image $K_e^{(n)}$

$$K_e^{(n)} = \left(S_e^{(n)} \odot DT\right) \cup \left(\overline{S_e^{(n)}} \odot L\right) \ , \tag{10}$$

where $S_e^{(n)}$ represents the binary image at step $n$, the subscript $e$ indicates that the rule for the external edge is being considered, and finally

$$L = \{l \in E^3 \mid l = (u, \lambda), \ \forall \, u \in E^2\} \ . \tag{11}$$

As shown in (Haralick et al., 1987), in order to compute the minimum value of a grey-tone image $K_e^{(n)}$ in a 4-connected neighborhood, the following grey-scale morphological erosion should be used:

$$M_e^{(n)} = K_e^{(n)} \ominus Q \ , \tag{12}$$

where

$$Q = \{(1, 0, 0); (-1, 0, 0); (0, 1, 0); (0, -1, 0)\} \ , \tag{13}$$

as shown in Figure 17.

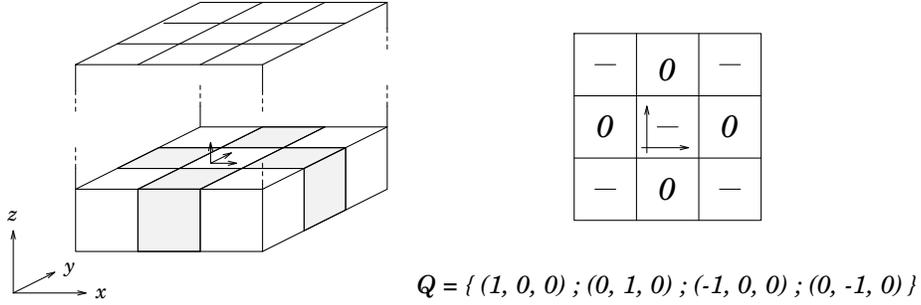

$$Q = \{ \, (1, 0, 0) \, ; (0, 1, 0) \, ; (-1, 0, 0) \, ; (0, -1, 0) \, \}$$

Figure 17: Structuring element $Q$

In order to determine the set of elements in which $M_e^{(n)}$ has a value smaller than $DT$, a new function $\mathcal{M}$ is required:

$$\mathcal{M} : E^3 \times E^3 \to E^2 \ . \tag{14}$$

Such a function is defined as

$$\mathcal{M}(A, B) \overset{\Delta}{=} \{x \in E^2 \mid \mathcal{V}(A, x) < \mathcal{V}(B, x)\} \ , \tag{15}$$

where $\mathcal{V} : E^3 \times E^2 \to E$ is defined as

$$\mathcal{V}(A, x) \overset{\Delta}{=} \begin{cases} a & \text{if } \exists \, a \in E \mid (x, a) \in T(A) \\ -\infty & \text{otherwise} \end{cases} \ . \tag{16}$$

In equation (16), $T(A)$ represents the *top* of $A$ (Haralick et al., 1987), here defined as

$$T(A) \overset{\Delta}{=} \{t \in A, \ t = (u, v) \mid \ \nexists \, t' = (x, v') \in A \text{ for which } v' > v\} \ . \tag{17}$$

The set of elements which will be included in set $S_e^{(n+1)}$ is given by the logical intersection between $\mathcal{M}(M_e^{(n)}, DT)$ and the set of elements belonging to the external edge of $S_e^{(n)}$:

$$E_e^{(n)} = \mathcal{M}\left(M_e^{(n)}, DT\right) \cap \mathcal{B}_e\left(S_e^{(n)}\right) \ . \tag{18}$$

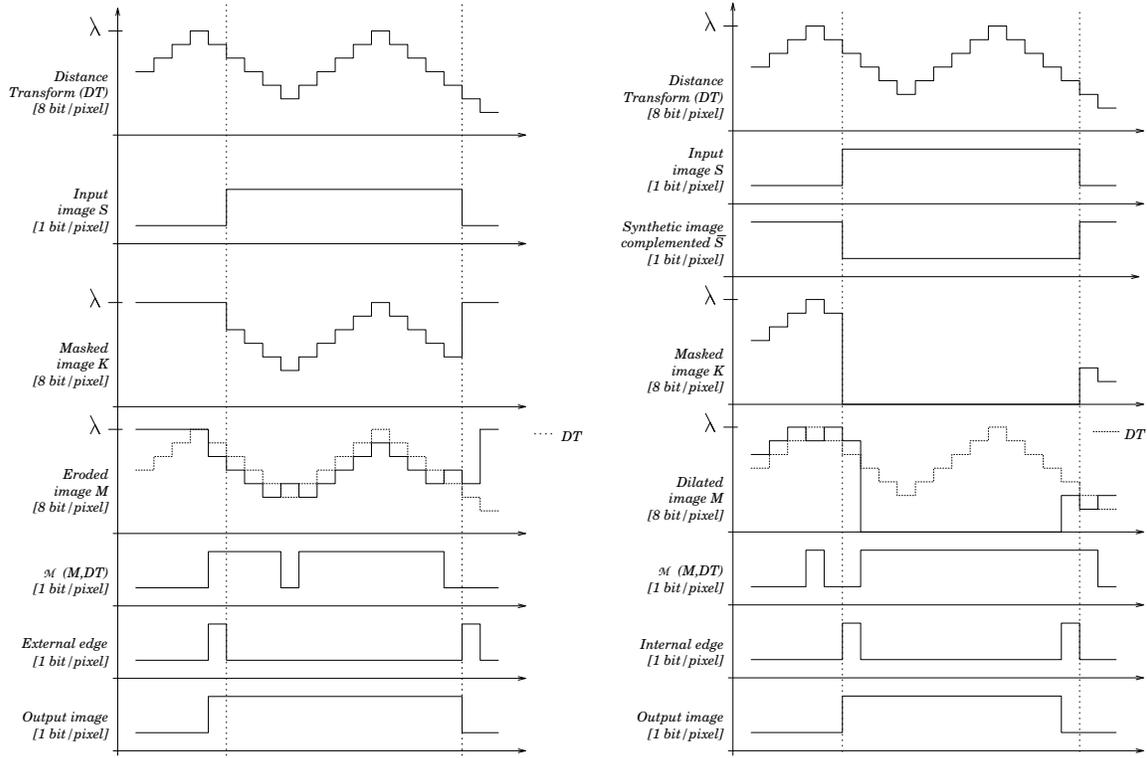

Figure 18: Monodimensional stretching in the case of external edge (left) and internal edge (right)

Thus, the final result of iteration $n$ is given by

$$S_e^{(n+1)} = S_e^{(n)} \cup E_e^{(n)} \ .\qquad(19)$$

Figure 18.a shows the execution of an individual iteration of rule 1 on a monodimensional image profile.

Following similar steps, it is possible to formalize rule 2. In order to compute the maximum value of the $DT$ in the specified neighborhood, let us consider image $K_i^{(n)}$

$$K_i^{(n)} = \overline{S_i^{(n)}} \odot DT \ ,\qquad(20)$$

where the subscript $i$ indicates that the rule for the internal edge is being considered. As shown above, in order to compute the maximum value of a grey-tone image $K_i^{(n)}$ in a 4-connected neighborhood, the following morphological dilation should be used:

$$M_i^{(n)} = K_i^{(n)} \oplus Q \ ,\qquad(21)$$

where $Q$ is shown in Figure 17.

The set of elements which will be removed from set $S_i^{(n+1)}$ is given by the logical intersection between $\mathcal{M}(M_i^{(n)}, DT)$ and the set of elements belonging to the internal edge of $S_i^{(n)}$:

$$E_i^{(n)} = \mathcal{M}\left(M_i^{(n)}, DT\right) \cap \mathcal{B}_i\left(S_i^{(n)}\right) \ .\qquad(22)$$

344

Thus, the final result of iteration $n$ is given by

$$S_i^{(n+1)} = \overline{\left(\overline{S_i^{(n)}} \cup E_i^{(n)}\right)} = S_i^{(n)} \cap \overline{E_i^{(n)}} \ . \tag{23}$$

Figure 18b shows the execution of an individual iteration of rule 2 on a monodimensional profile of an image.

## Acknowledgements

This work was partially supported by the Italian CNR within the framework of the Eureka PROMETHEUS Project – Progetto Finalizzato Trasporti under contracts n. 93.01813.PF74 and 94.01371.PF74.

The authors are indebted to Gianni Conte for the valuable and constructive discussions and for his continuous support throughout the project.